\def\year{2018}\relax
\documentclass[letterpaper]{article} 
\usepackage{aaai18}  
\usepackage{times}  
\usepackage{helvet}  
\usepackage{courier}  
\usepackage{url}  
\usepackage{graphicx}  
\usepackage{amsmath}
\usepackage{changepage}
\begin{document}
%

\nocopyright

\title{Event Representations for Automated Story Generation with Deep Neural Nets}
\author{}
\author{Lara J. Martin, Prithviraj Ammanabrolu, Xinyu Wang,\AND William Hancock, Shruti Singh, Brent Harrison, and Mark O. Riedl\\
School of Interactive Computing\\
	Georgia Institute of Technology\\
	Atlanta, GA\\
    \textit{[ljmartin; raj.ammanabrolu; lillywang1126; whancock; shruti.singh; brent.harrison; riedl]@gatech.edu}
}
\maketitle
\begin{abstract}
Automated story generation is the problem of automatically selecting a sequence of events, actions, or words that can be told as a story. We seek to develop a system that can generate stories by learning everything it needs to know from textual story corpora. To date, recurrent neural networks that learn language models at character, word, or sentence levels have had little success generating coherent stories.
We explore the question of event representations that provide a mid-level of abstraction between words and sentences in order to retain the semantic information of the original data while minimizing event sparsity. We present a technique for preprocessing textual story data into event sequences. We then present a technique for automated story generation whereby we decompose the problem into the generation of successive events ({\em event2event}) and the generation of natural language sentences from events ({\em event2sentence}). We give empirical results comparing different event representations and their effects on event successor generation and the translation of events to natural language.
\end{abstract}

\section{Introduction}

{\em Automated story generation} is the problem of automatically selecting a sequence of events, actions, or words that can be told as a story.
To date, most story generation systems have used symbolic planning ~\cite{meehan76,lebowitz87,perez01,Porteous2009,riedl:jair2010} or case-based reasoning ~\cite{gervas05}.
While these automated story generation systems were able to produce impressive results, they rely on a human-knowledge engineer to provide symbolic domain models that indicated legal characters, actions, and knowledge about when character actions can and cannot be performed; 
these systems are limited to only telling stories about topics that are covered by the domain knowledge.
Consequently, it is difficult to determine whether the quality of the stories produced by these systems is a result of the algorithm or good knowledge engineering. 

{\em Open story generation} ~\cite{Li2013} is the problem of automatically generating a story about any topic without {\it a priori} manual domain knowledge engineering. 
Open story generation requires an intelligent system to either learn a domain model from available data ~\cite{Li2013,roemmele2017} or to reuse data and knowledge available from a corpus \cite{swanson12}.

In this paper, we explore the use of recurrent encoder-decoder neural networks (e.g., {\em Sequence2Sequence} ~\cite{sutskever2014}) for open story generation. 
An encoder-decoder RNN is trained to predict the next token(s) in a sequence, given one or more input tokens. The network architecture and set of weights $\theta$ comprise a generative model capturing and generalizing over patters observed in the training data.
For open story generation, we must train the network on a dataset that encompasses as many story topics as possible.
For this work, we use a corpus of movie plot summaries extracted from Wikipedia ~\cite{bamman2014} under the premise that the set of movies plots on Wikipedia covers the range of topics that people want to tell stories about.

In narratological terms, an {\em event} is a unit of story featuring a world state change \cite{prince87}. Textual story corpora, including Wikipedia movie plot corpora, is comprised of unstructured textual sentences. One benefit to dealing with movie plots is its clarity of events that occur, although this is often to the expense of more creative language.
Even so, character- or word-level analysis of these sentences would fail to capture the interplay between the words that make up the meaning behind the sentence. 
Character- and word-level recurrent neural networks can learn to create grammatically correct sentences but often fail to produce coherent narratives beyond a couple of sentences.
On the other hand, sentence-level events would be too unique from each other to find any real relationship between them.
Even with a large corpus of stories, we would most likely have sequences of sentences that would only ever be seen once. For example, ``Old ranch-hand Frank Osorio travels from Patagonia to Buenos Aires to bring the news of his daughter's demise to his granddaughter Alina.'' occurs only once in our corpus, so we have only ever seen one example of what is likely to occur before and after it (if anything).
Due to event sparsity, we are likely to have poor predictive ability.

In order to help maintain a coherent story, one can provide an event representation that is expressive enough to preserve the semantic meaning of sentences in a story corpus while also reducing the sparsity of events (i.e. increasing the potential overlap of events across stories and the number of examples of events the learner observes).
In this paper, we have developed an event representation that aids in the process of automated, open story generation. 
The insight is that if one can extract some basic semantic information from the sentences of preexisting stories, one can learn the skeletons of what ``good'' stories are supposed to be like. 
Then, using these templates, the system will be able to generate novel sequences of events that would resemble a decent story.

The first contribution of our paper is thus an event representation and a proposed recurrent encoder-decoder neural network for story generation called {\em event2event}. We evaluate our event representation against the naive baseline sentence representation and a number of alternative representations.

In {\em event2event}, a textual story corpus is preprocessed---sentences are translated into our event representation by extracting the core semantic information from each sentence.
Event preprocessing is a linear-time algorithm using a number of natural language processing techniques. 
The processed text is then used to train the neural network.
However, event preprocessing is a lossy process and the resultant events are not human-readable.
To address this, we present a story generation pipeline in which a second neural network, {\em event2sentence}, translates abstract events back into natural language sentences. 
The {\em event2sentence} network is an encoder-decoder network trained to fill in the missing details necessary for the abstract events to be human-readable.

\begin{sloppypar}
Our second contribution is the overall story generation pipeline in which subsequent events of a story are generated via an {\em event2event} network and then translated into natural language using an {\em event2sentence} network. We present an evaluation of {\em event2sentence} on different event representations and draw conclusions about the effect of representations on the ability to produce readable stories.
%
%
\end{sloppypar}

The remainder of the paper is organized as follows. 
First, we discuss related work on automated story generation, followed by an introduction of our event representation. Then we introduce our {\em event2event} network and provide an evaluation of the event representation in the context of story generation.
We show how the event representation can be used in {\em event2sentence} to generate a human-readable sentences from events. We end with a discussion of future work and conclusions about these experiments and how our event representation and event-to-sentence model will fit into our final system.


\section{Related Work}

Automated Story Generation has been a research problem of interest since nearly the inception of artificial intelligence. 
Early attempts relied on symbolic planning ~\cite{meehan76,lebowitz87,perez01,riedl:jair2010} or case-based reasoning using ontologies ~\cite{gervas05}.
These techniques could only generate stories for predetermined and well-defined domains of characters, places, and actions. 
The creativity of these systems conflated the robustness of manually-engineered knowledge and algorithm suitability.

Recently, machine learning has been used to attempt to learn the domain model from which stories can be created or to identify segments of story content available in an existing repository to assemble stories. 
The {\em SayAnthing} system \cite{swanson12} uses textual case-based reasoning to identify relevant existing story content in online blogs.
The {\em Scheherazade} system \cite{Li2013} uses a crowdsourced corpus of example stories to learn a domain model from which to generate novel stories.

Recurrent neural networks can theoretically learn to predict the probability of the next character, word, or sentence in a story. 
Roemmele and Gordon \cite{roemmele2017} use a Long Short-Term Memory (LSTM) network ~\cite{lstm} to generate stories. 
They use Skip-thought vectors~\cite{kiros2015} to encode sentences and a technique similar to word2vec ~\cite{word2vec} to embedded entire sentences into 4,800-dimensional space.
They trained their network on the BookCorpus dataset.
Khalifa et al. \shortcite{khalifa2017} argue that stories are better generated using recurrent neural networks trained on highly specialized textual corpora, such as the body of works from a single, prolific author.
However, such a technique is not capable of {\em open story generation}.

Based off of the theory of script learning ~\cite{Schank1977}, Chambers and Jurafsky  \shortcite{chambers2008} learn causal chains that revolve around a protagonist. They developed a representation that took note of the event/verb that occurred and the type of dependency that connected the event to the protagonist (e.g. was the protagonist the object of this event?).

Pichotta and Mooney \shortcite{Pichotta2016} developed a 5-tuple event representation of $(v,e_s,e_o,e_p,p)$, where $v$ is the verb, $p$ is a preposition, and  $e_s$, $e_o$, and $e_p$ are nouns representing the subject, direction object, and prepositional object, respectively. 
Our representation was inspired by this work, although we use a slightly different representation.
Because it was a paper on script learning, they did not need to convert the event representations back into natural language.

Related to automated story generation, the {\em story cloze test} \cite{mostafazadeh2016} is the task of choosing between two given endings to a story.
The story cloze test transforms story generation into a classification problem:
a 4-sentence story is given along with two alternative sentences that can be the 5th sentence.
State-of-the art story cloze test techniques use a combination of word embeddings, sentiment analysis, and stylistic features \cite{mostafazadeh2017}.


\section{Event Representation}

Automated story generation can be formalized as follows: given a sequence of events, sample from the probability distribution over successor events. 
That is, simple automated story generation can be expressed as a process whereby the next event is computed by sampling or maximizing 
$Pr_\theta(e_{t+1}|e_{t-k},...,e_{t-1},e_{t})$
where $\theta$ is the set of parameters of a generative domain model, $e_i$ is the event at time $i$, and where $k$ indicates the size of a sliding window of context, or history.

In our work, the probability distribution is produced by a recurrent encoder-decoder network with parameters $\theta$.
In this section, we consider what the level of abstraction for the inputs into the network should be such that it produces the best predictive power while retaining semantic knowledge.
Event sparsity results in a situation where all event successors have a low probability of occurrence, potentially within a margin of error. 
In this situation, story generation devolves to a random generation process.

Following Pichotta and Mooney ~\shortcite{Pichotta2016}, we developed a 4-tuple event representation $\langle s, v, o, m\rangle$ where $v$ is a verb, $s$ is the subject of the verb, $o$ is the object of the verb, and $m$ is the modifier or ``wildcard'', which can be a propositional object, indirect object, causal complement (e.g., in ``I was glad that he drove,'' ``drove'' is the causal complement to ``glad.''), or any other dependency unclassifiable to Stanford's dependency parser. All words were stemmed. 
Events are created by first extracting dependencies with Stanford's CoreNLP ~\cite{manning2014} and locating the appropriate dependencies mentioned above. 
If the object or modifier cannot be identified, we insert the placeholder $EmptyParameter$, which we will refer to as $\emptyset$ in this paper.

\begin{sloppypar}
Our event translation process can either extract a single event from a sentence or multiple events per sentence. 
If we were to extract multiple events, it is because there are verbal or sentential conjunctions in the sentence.
Consider the sentence ``John and Mary went to the store,'' our algorithm would extract two events: $\langle john, go, store, \emptyset\rangle$ and $\langle mary, go, store, \emptyset\rangle$. 
The average number of events per sentence was 2.69.
\end{sloppypar}

\begin{sloppypar}
Our experiments below used a corpus of movie plots from Wikipedia~\cite{bamman2014}, which we cleaned to any remove extraneous Wikipedia syntax, such as links for which actors played which characters. This corpus contains 42,170 stories with the average number of sentences per story being 14.515.
\end{sloppypar}

The simplest form of our event representation is achieved by extracting the verb, subject, object, and modifier term from each sentence. 
However, there are variations on the event representation that increase the level of abstraction (and thus decrease sparsity) and help the encoder-decoder network predict successor events. We enumerate some of the possible variations below.

\begin{itemize}
\item
{\bf Generalized.}
Each element in the event tuple undergoes further abstraction.
Named entities were identified (cf. \cite{finkel2005}), and ``PERSON'' names were replaced with the tag \textless NE\textgreater $n$, where $n$ indicates the $n$-th named entity in the sentence. Other named entities were labeled as their NER category (e.g. LOCATION, ORGANIZATION, etc.).
The rest of the nouns were replaced by the WordNet ~\cite{wordnet} synset two levels up in the inherited hypernym hierarchy, giving us a general category (e.g. self-propelled\_vehicle$.n.01$ vs the original word ``car'' (car$.n.01$)), while avoiding labeling it too generally (e.g. entity$.n.01$). Verbs were replaced by VerbNet ~\cite{verbnet} version 3.2.4~\footnote{https://verbs.colorado.edu/vn3.2.4-test-uvi/index.php} frames (e.g. ``arrived'' becomes ``escape-$51.1$'', ``transferring'' becomes ``contribute-$13.2$-2'').

\item
{\bf Named Entity Numbering.}
There were two ways of numbering the named entities (i.e. people's names) that we experimented with. One way had the named entity numbering reset with every sentence (consistent within sentence)--or, {\em sentence NEs}, our ``default''. The other way had the numbering reset after every input-output pair (i.e. every line of data; consistent across two sentences)--or, {\em continued NEs}.

\item
{\bf Adding Genre Information.}
We did topic modeling on the entire corpus using Python's Latent Dirichlet Analysis~\footnote{https://pypi.python.org/pypi/lda} set for discovering 100 different categories. We took this categorization as a type of emergent genre classification. Some clusters had a clear pattern, e.g., ``job company work money business''. Others were less clear. Each cluster was given a unique genre number which was added to the event representation to create a 5-tuple $\langle s, v, o, m, g\rangle$ where $s$, $v$, $o$, and $m$ are defined as above and $g$ is the genre cluster number.

\end{itemize}

We note that other event representations can exist, including representations that incorporate more information as in \cite{Pichotta2016}. 
The experiments in the next section show how different representations affect the ability of a recurrent neural network to predict story continuations.


\section{Event-to-Event}

\begin{sloppypar}
The {\em event2event} network is a recurrent multi-layer encoder-decoder network based on \cite{sutskever2014}.
Unless otherwise stated in experiments below, our {\em event2event} network is trained with input $x=w^{n}_1,w^{n}_2,w^{n}_3,w^{n}_4$ where each $w^{n}_i$ is either $s$, $v$, $o$, or $m$ from the $n$-th event, and output $y=w^{n+1}_1,w^{n+1}_2,w^{n+1}_3,w^{n+1}_4$.
\end{sloppypar}
The experiments described below seek to determine how different event representations affected {\em event2event} predictions of the successor event in a story.
We evaluated each event representation using two metrics. 
{\em Perplexity} is the measure of how ``surprised'' a model is by a training set.
Here we use it to gain a sense of how well the probabilistic model we have trained can predict the data. Specifically, we built the model using an $n$-gram length of $1$:

\begin{equation}
  Perplexity = 2^{-\sum_{x}p(x)\log_2{p(x)}}
\end{equation}
\noindent
where $x$ is a token in the text, and
\begin{equation}
  p(x) = \frac{count(x)}{\sum_{x}count(x)}
\end{equation}

\noindent 
The larger the unigram perplexity, the less likely a model is to produce the next unigram in a test dataset. 

The second metric is BLEU score, which compares the similarity between the generated output and the ``ground truth'' by looking at $n$-gram precision.
The neural network architecture we use was initially envisioned for machine translation purposes, where BLEU is a common evaluation metric. 
Specifically, we use an $n$-gram length of $4$ and so the score takes into account all $n$-gram overlaps between the generated and expected output where $n$ varies from $1$ to $4$ ~\cite{papineni2002}.

We use a greedy decoder to produce the final sequence by taking the token with the highest probability at each step.
\begin{equation}
\hat{W}=\operatorname*{arg\,max}_w Pr(w|S)
\end{equation}
\noindent
where $\hat{W}$ is the generated token appended to the hypothesis, $S$ is the input sequence, and $w$  represents the possible output tokens.

\subsection{Experimental Setup} \label{e2e:setup}
For each experiment, we trained a sequence-to-sequence recurrent neural net~\cite{sutskever2014} using Tensorflow ~\cite{tensorflow}. Each network was trained with the same parameters (0.5 learning rate, 0.99 learning rate decay, 5.0 maximum gradient, 64 batch size, 1024 model layer size, and 4 layers), varying only the input/output, the bucket size, the number of epochs and the vocabulary. The neural nets were trained until the decrease in overall loss was less than $5\%$ per epoch. This took between $40$ to $60$ epochs for all experiments.
The data was split into $80\%$ training, $10\%$ validation, and $10\%$ test data. All reported results were evaluated using the the held-out test data.

We evaluated 11 versions of our event representation against a sentence-level baseline.
Numbers below correspond to rows in results Table~\ref{table:e2e}.

\begin{enumerate}
\setcounter{enumi}{-1}

\item {\em Original Sentences.} As our baseline, we evaluated how well an original sentence can predict its following original sentence within a story.

\begin{sloppypar}
\item {\em Original Words Baseline.} We took the most basic, 4-word event representation: $\langle s, v, o, m\rangle$ with no abstraction and using original named entity names.
\end{sloppypar}

\item {\em Original Words with \textless NE\textgreater s.} Identical to the previous experiment except entity names that were classified as ``PERSON'' through NER were substituted with \textless NE\textgreater$n$. 

\item {\em Generalized.} The same 4-word event structure except with named entities replaced and all other words generalized through WordNet or VerbNet, following the procedure described earlier.

\end{enumerate}

To avoid an overwhelming number of experiments, the next set of experiments used the ``winner'' of the first set of experiments.
Subsequent experiments used variations of the generalized event representation (\#3), which showed drastically lower perplexity scores.

\begin{enumerate}
\setcounter{enumi}{3}

 \item {\em Generalized, Continued \textless NE\textgreater s.} This experiment mirrors the previous with the exception of the number of the \textless NE\textgreater s. In the previous experiment, the numbers restarted after every event. Here, the numbers continue across input and output. So if $event_1$ mentioned ``Kendall'' and $event_2$ (which follows $event_1$ in the story) mentioned ``Kendall'', then both would have the same number for this character.

\item {\em Generalized + Genre.} This is the same event structure as experiment \#3 with the exception of an additional, $5^{th}$ parameter in the event: genre. The genre number was used in training for {\em event2event} but removed from inputs and outputs before testing; it artificially inflated BLEU scores because it was easy for the network to guess the genre number as the genre number was weighted equally to other words.

\item {\em Generalized Bigram.} This experiment tests whether RNN history aids in predicting the next event. We modified {\em event2event} to give it the event bigram $e_{n-1}, e_n$ and to predict $e_{n+1}, e_{n+2}$. We believe that this experiment could generalize to cases with a $e_{n-k},...,e_{n}$ history.

\item {\em Generalized Bigram, Continued \textless NE\textgreater s.} This experiment has the same continued NE numbering as experiment \#4 had but we trained {\em event2event} with event bigrams.

\item {\em Generalized Bigram + Genre.} This is a combination of the ideas from experiments \#5 and \#6: generalized events in event bigrams and with genre added.

\end{enumerate}

The following three experiments investigate extracting more than one event per sentence in the story corpus when possible; the prior experiments only use the first event per sentence in the original corpus.

\begin{enumerate}
\setcounter{enumi}{8}

\item {\em Generalized Multiple, Sequential.} 
When a sentence yields more than one event, $e^{1}_n, e^{2}_n,...$ where $n$ is the $n$th sentence and $e^{i}_n$ is the $i$th event created from the $n$th sentence, we train the neural network as if each event occurs in sequence, i.e., $e^{1}_n$ predicts $e^{2}_n$, $e^{2}_n$ predicts $e^{3}_n$, etc.
The last event from sentence $n$ predicts the first event from sentence $n+1$. 

\item {\em Generalized Multiple, Any Order.}
Here we gave the RNN all orderings of the events produced by a single sentence paired, in turn, with all orderings of each event of the following sentence.

\item {\em Generalized Multiple, All to All.}
In this experiment, we took all of the events produced by a single sentence together as the input, with all of the events produced by its following sentence together as output. For example, if sentence $i$ produced events $e^{1}_i$, $e^{2}_i$, and $e^{3}_i$, and the following sentence $j$ produced events $e^{1}_j$ and $e^{2}_j$, then we would train our neural network on the input: $e^{1}_i$ $e^{2}_i$ $e^{3}_i$, and the output: $e^{1}_j$ $e^{2}_j$.

\end{enumerate}

\subsection{Results and Discussion}

The results from the experiments outlined above can be found in Table ~\ref{table:e2e}.

\begin{table}[b]
\footnotesize
  \caption{\footnotesize Results from the event-to-event experiments. Best values from each of these three sections (baselines, additions, and multiple events) are bolded.}
  \label{table:e2e}
  \begin{tabular}{ | l | c | c | }
    \hline
    \em{\bf Experiment} & \em{\bf Perplexity} & \em{\bf BLEU}\\
    \hline
    (0) Original Sentences & 704.815 & 0.0432\\
    \hline
    (1) Original Words Baseline & 748.914 & \textbf{0.1880}\\
    \hline
    (2) Original Words with \textless NE\textgreater s & 166.646 & 0.1878\\
    \hline
    (3) Generalized Baseline & \textbf{54.231} & 0.0575\\
    \hline\hline
    (4) Generalized, Continued NEs & 56.180 & 0.0544\\
     \hline
    (5) Generalized + Genre & \textbf{48.041} & 0.0525\\
    \hline
    (6) Generalized Bigram & 50.636 & 0.1549\\
    \hline
    (7) Generalized Bigram, Cont. NEs & 50.189 & \textbf{0.1567}\\
    \hline
    (8) Generalized Bigram + Genre & 48.505 & 0.1102\\
    \hline\hline
    (9) Generalized Multiple, Sequential & 58.562 & 0.0521\\
    \hline
    (10) Generalized Multiple, Any Order & 61.532 & 0.0405\\
    \hline
    (11) Generalized Multiple, All to All & \textbf{45.223} & \textbf{0.1091}\\
    \hline
  \end{tabular}
\end{table}

The original word events had similar perplexity to original sentences.
This parallels similar observations made by Pichotta and Mooney \cite{Pichotta2016b}.
Deleting words did little to improve the predictive ability of our {\em event2event} network.
However, perplexity improved significantly once character names were replaced by generalized \textless NE\textgreater tags, followed by generalizing other words and verbs. 

Overall, the generalized events had much better perplexity scores, and making them into bigrams---incorporating history---improved the BLEU scores to nearly those of the original word events. Adding in genre information improved perplexity.

The best perplexity was achieved when multiple generalized events were created from sentences as long as all of the events were fed in at the same time (i.e. no order was being forced upon the events that came from the same sentence). The training data was set up to encourage the neural network to correlate all of the events in one sentence with all of the events from the next sentence.

Although the events with the original words (with or without character names) performed better in terms of BLEU score, it is our belief that BLEU is not the most appropriate metric for event generation because it emphasizes the recreation of the input.
Overall, BLEU scores are very low for all experiments, attesting to the inappropriateness of the metric.
Perplexity is a more appropriate metric for event generation because it correlates with the ability for a model to predict the entire test dataset.
Borrowing heavily from the field of language modeling, the recurrent neural network approach to story generation is a prediction problem.

Our intuition that the generalized events would perform better in generating successive events bears out in the data. 
However, greater generalization makes it harder to return events to natural language sentences.
We also see that the BLEU scores for the bigram experiments are generally higher than the others. This shows that history matters and that the additional context provided increases the number of $n$-gram overlaps between the generated and expected outputs.

The movie plots corpus contains numerous sentences that can be interpreted as describing multiple events.
Naive implementation of multiple events hurt perplexity because there is no implicit order of events generated from the same sentence; they are not necessarily sequential. 
When we allow multiple events from sentences to be followed by all of the events from a subsequent sentence, perplexity improves.


\section{Event-to-Sentence}

Unfortunately, events are not human-readable and must be converted to natural language sentences.
Since the conversion from sentences to (multiple) events for {\em event2event} is a linear and lossy process, the translation of events back to sentences is non-trivial as it requires adding details back in.
For example, the event $\langle$relative.n.01, characterize-29.2, male.n.02, feeling.n.01$\rangle$ could, hypothetically, have come from the sentence ``Her brother praised the boy for his empathy.'' In actuality, this event came from the sentence ``His uncle however regards him with disgust.''

Complicating the situation, the {\em event2event} encoder-decoder network is not guaranteed to produce an event that has ever been seen in the training story corpus.
Furthermore, our experiments with event representations for {\em event2event} indicate that greater generalization lends to better story generation. 
However, the greater the generalization, the harder it is to translate an event back into a natural language sentence.

In this section we introduce {\em event2sentence}, a neural network designed to translate an event into natural language.
The {\em event2event} network takes an input event $e_n=\langle s^n, v^n, o^n, v^n\rangle$ and samples from a distribution over possible successor events $e_{n+1}=\langle s^{n+1}, v^{n+1}, o^{n+1}, m^{n+1}\rangle$. 
As before, we use a recurrent encoder-decoder network based on \cite{sutskever2014}.
The {\em event2sentence} network is trained on parallel corpora of sentences from a story corpus and the corresponding events. 
In that sense, {\em event2sentence} is attempting to learn to reverse the lossy event creation process.

\begin{figure}[tb]
\centering
\includegraphics[width=2.5in]{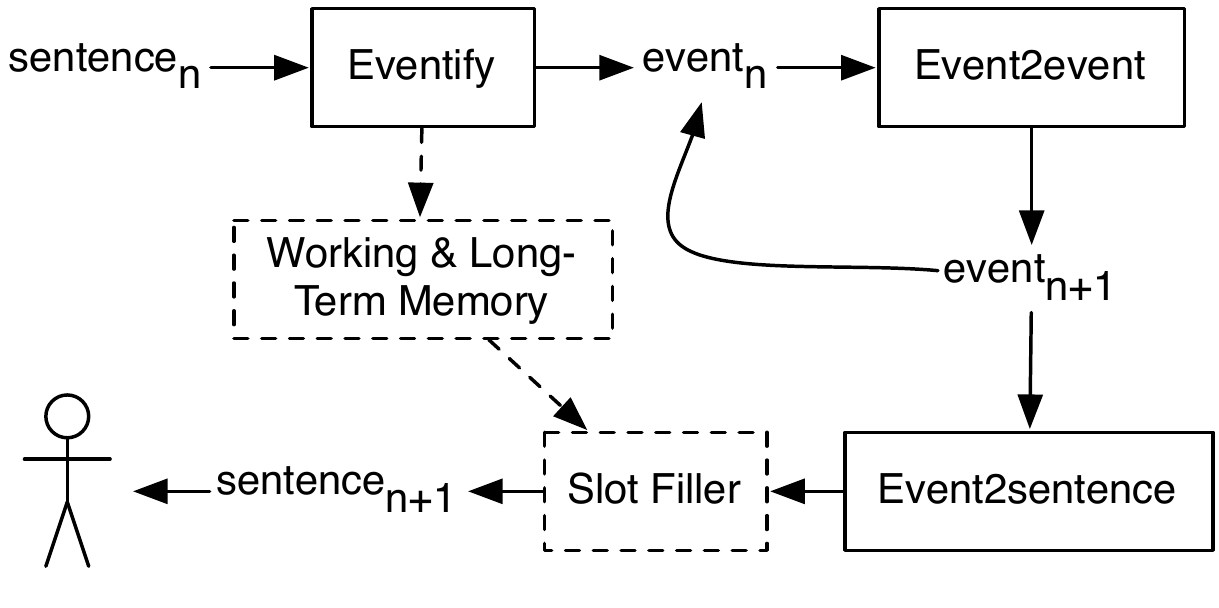} 
\caption{\footnotesize Our automated story generation pipeline. Dashed boxes and arrows represent future work.}
\label{fig:pipeline}
\end{figure}

We envision {\em event2event} and {\em event2sentence} working together as illustrated in Figure \ref{fig:pipeline}. First, a sentence---provided by a human---is turned into one or more events. The {\em event2event} network generates one or more successive events. The {\em event2sentence} network translates the events back into natural language and presents it to the human reader.
The dashed lines and boxes represent future work for filling in story specifics.
To continue story generation, $event_{n+1}$ can be fed back into {\em event2event}; the sentence generation is purely for human consumption. 

The {\em event2sentence} experiments in the next section investigate how well different event representations can be ``translated'' back into natural language sentences.

\subsection{Experimental Setup}

The setup for this set of experiments is almost identical to that of the {\em event2event} experiments, with the main difference being that we used PyTorch~\footnote{http://pytorch.org/} which more easily lent itself to implementing beam search. The LSTM RNN networks in these experiments use beam search instead of greedy search to aid in finding a more optimal solution while decoding.



The beam search decoder works by maintaining a number of partial hypotheses at each step (known as the beam width or $B$, where $B$=5). Each of these hypotheses is a potential prefix of a sentence. At each step, the $B$ tokens with the highest probabilities in the distribution are used to expand the partial hypotheses. This continues until the end-of-sentence tag is reached.

%
The input for these experiments was the events of a particular representation and the output was a newly-generated sentence based on the input event(s). The models in these experiments were trained on the events paired with the sentences they were ``eventified'' from. In a complete story generation system, the output of the {\em event2event} network feeds into the {\em event2sentence} network. Examples of this can be seen in Table~\ref{table:examples}. However, we tested the {\em event2sentence} network on the same events extracted from the original sentences as were used for {\em event2event} in order to conduct controlled experiments and compute perplexity and BLEU scores.

To test {\em event2sentence} with an event representation that used the original words is relatively straight forward.
Experimenting on translating generalized events to natural language sentences was more challenging since we would be forcing the neural net to guess character names, nouns, and verbs.

\begin{sloppypar}
We devised an alternative approach for generalized {\em event2sentence} whereby sentences were first partially eventified.
That is, we trained {\em event2sentence} on generalized sentences where the ``PERSON'' named entities were replaced by \textless NE\textgreater tags, other named entities were replaced by their NER category, and the remaining nouns were replaced with WordNet synsets. The verbs were left alone since they often do not have to be consistent across sentences within a story.
The intuition here is that the character names and particulars of objects and places are highly mutable and do not affect the overall flow of a story as long as they remain consistent. 
\end{sloppypar}

Below, we show an example of a sentence and its partially generalized counterpart. The original sentence

\begin{quote}
\footnotesize
The remaining craft launches a Buzz droid at the ARC 1 7 0 which lands near the Clone Trooper rear gunner who uses a can of Buzz Spray to dislodge the robot.
\end{quote}
would be partially generalized to 
\begin{quote}
\begin{sloppypar}
\footnotesize
The remaining activity$.n.01$ launches a happening$.n.01$ droid at the ORGANIZATION 1 7 0 which property$.n.01$ near the person$.n.01$ enlisted\_person$.n.01$ rear skilled\_worker$.n.01$ who uses a instrumentality$.n.03$ of happening$.n.01$ chemical$.n.01$ to dislodge the device$.n.01$
\end{sloppypar}
\end{quote}

We also looked at whether {\em event2sentence} performance would be improved if we used multiple events per sentence (when possible) instead of the default single event per sentence.
Alternatively, we automatically split and prune (S+P) sentences; removing prepositional phrases, splitting sentential phrases on conjunctions, and, when it does not start with a pronoun (e.g. who), splitting S' (read: S-bar) from its original sentence and removing the first word.
This would allow us to evaluate sentences that would have fewer (ideally one) events extracted from each. For example, 
\begin{quote}
\footnotesize
Lenny begins to walk away but Sam insults him so he turns and fires, but the gun explodes in his hand.
\end{quote}
becomes
\begin{quote}
\footnotesize
Lenny begins to walk away. Sam insults him. He turns and fires. The gun explodes.
\end{quote}
Although splitting and pruning the sentences should bring most sentences down to a single event, this isn't always the case. Thus, we ran an {\em event2sentence} experiment where we extracted all of the events from the S+P sentences.


\subsection{Results and Discussion}
The results of our {\em event2sentence} experiments are shown in Table~\ref{table:e2s}.
Although generalizing sentences improves perplexity drastically, splitting and pruning sentences yields better BLEU scores when the original words are kept. 
In the case of {\em event2sentence}, BLEU scores make more sense as a metric since the task is a translation task. Perplexity in these experiments appears to correspond to vocabulary size.

Generalized events with full-length generalized sentences have better BLEU scores than when the original words are used. However, when we work with S+P sentences, the pattern flips. We believe that because both S+P and word generalizing methods reduce sparsity of events, when they are combined too much information is lost and the neural network struggles to find any distinguishing patterns.

\begin{table}[t]
\footnotesize
  \caption{\footnotesize Results from the event-to-sentence experiments.}
  \label{table:e2s}
  \begin{tabular}{ | p{5.3cm} | p{1.2cm} | p{0.8cm} | }
    \hline
    \em{\bf Experiment} & \centering{\em{\bf Perplexity}} & \em{\bf BLEU}\\
    \hline
    Original Words Event $\rightarrow$ Original Sentence & \centering{1585.46} & 0.0016 \\
    \hline
    Generalized Event $\rightarrow$ Generalized Sentence & \centering{56.516} & 0.0331 \\
    \hline
    All Generalized Events $\rightarrow$ Gen. Sentence & \centering{59.106} & 0.0366\\
    \hline\hline
    Original Words Event $\rightarrow$ S+P Sentence & \centering{490.010} & \textbf{0.0764}\\
    \hline
    Generalized Event $\rightarrow$ S+P Gen. Sentence &  \centering{\textbf{53.964}} &  0.0266\\
    \hline
    All Generalized Events $\rightarrow$ S+P Gen. Sent. &  \centering{56.488} &  0.0283\\
    \hline
  \end{tabular}
\end{table}

\begin{table*}[t]
\footnotesize
  \caption{\footnotesize End-to-End Examples. Let $\emptyset$ represent an empty (unfilled) parameter. Commas are added for readability.}
  \label{table:examples}

  \begin{tabular}{ | p{2cm}| p{3cm} | p{3cm} | p{4.2cm} | p{3.4cm} |}
    \hline
    \em{\bf Experiment} & \em{\bf Input (Previously Unseen)} & \em{\bf Extracted Event(s)} & \em{\bf Generated Next Event(s)} & \em{\bf Generated Next Sentence} \\
    \hline
   All Generalized Events \& Generalized Sentence &
   He reaches out to Remus Lupin, a Defence Against the Dark Arts teacher who is eventually revealed to be a werewolf. &
   $\langle$male$.n.02$ get-13.5.1 $\emptyset$ \textless NE\textgreater 0$\rangle$, $\langle$ORGANIZATION say-37.7-1 monster$.n.01$  $\emptyset\rangle$   &
   $\langle$monster$.n.01$ amuse-31.1 sarge $\emptyset\rangle$, $\langle$monster$.n.01$ amuse-31.1 realize $\emptyset\rangle$, $\langle$monster$.n.01$ conjecture-29.5-1 $\emptyset$ $\emptyset\rangle$, $\langle$male$.n.02$ conduit$.n.01$ entity$.n.01$ $\emptyset\rangle$, $\langle$male$.n.02$ free-80-1 $\emptyset$ penal\_institution$.n.01\rangle$ &
   When monster$.n.01$ nemesis$.n.01$ describes who finally realizes $male.n.02$ can not, dangerous entity$.n.01$ male$.n.02$ is released from penal\_institution$.n.01$. \\
\hline
    
     Generalized Event \& Generalized S+P Sentence &
     John unwittingly unleashes an insidious pox that rapidly spreads across the Caribbean. &
     $\langle$\textless NE\textgreater 0 disassemble-23.3 contagious\_disease$.n.01$ $\emptyset\rangle$ &
     $\langle$male$.n.02$ spatial\_configuration-47.6 $\emptyset$ adopt-93 $\rangle$ &
 male$.n.02$ crumples and is about to be sheath$.n.02$. \\
    \hline

     Original Words Event \& S+P Sentence &
     He thinks he's on a top secret spying mission. &
     $\langle$he think $\emptyset$ $\emptyset\rangle$ &
     $\langle$she come know truth$\rangle$ &
 	 She has come to the truth. \\
    \hline
    
  \end{tabular}
\end{table*}


Table~\ref{table:examples} shows examples from the entire pipeline as it currently exists, that is from one sentence to the next sentence without slot filling (See Figure~\ref{fig:pipeline}).
To get a full sense of how the generalized sentences would read, imagine adding character names and other details as if one were completing a {\em Mad-Libs} game.


\section{Future Work}
The question remains how to determine exactly what character names and noun details to use in place for the \textless NE\textgreater s and WordNet placeholders.
In Figure~\ref{fig:pipeline}, we propose the addition of Working Memory and Long-Term Memory modules. 
The Working Memory module would retain the character names and nouns in a lookup table that were removed during the eventification process. 
After a partially generalized sentence is produced by {\em event2sentence}, the system can use the Working Memory lookup table to fill character names and nouns back into the placeholders. 
The intuition is that from one event to the next, many of the details---especially character names---are likely to be reused.

In stories it is common to see a form of turn-taking between characters. 
For example the two events ``John hits Andrew'' \& ``Andrew runs away from John'' followed by ``John chases Andrew'' illustrate the turn-taking pattern.
If John was always used as the first named entity, the meaning of the example would be significantly altered.
The continuous numbering of named entities ({\em event2event} experiment \#7) is designed to assist event bigrams with maintaining turn-taking patterns.

There are times when the Working Memory will not be able to fill named entity and WordNet synset placeholder slots because the most recent event bigram does not contain the element necessary for reuse. 
The Long-Term Memory maintains a history of all named entities and nouns that have ever been used in the story and information about how long ago they were last used.
See \cite{Martin2016} for a cognitively-plausible event-based memory that can be used to compute the salience of entities in a story.
The underlying assumption is that stories are more likely to reuse existing entities and concepts than introduce new entities and concepts.

Our model of automated story generation as prediction of successor events is simplistic;
it assumes that stories can be generated by a language model that captures generalized patterns of event co-occurrence.
Story generation can also be formalized as a planning problem, taking into account communicative goals. 
In storytelling, a communicative goal can be to tell a story about a particular topic, to include a theme, or to end the story in a particular way. 
In future work, we plan to replace the {\em event2event} network with a reinforcement learning process that can perform lookahead to analyze whether potential successor events are likely to lead to communicative intent being met.

\section{Conclusions}
In automated story generation, event representation matters.
We hypothesize that by using our intuitions into storytelling we can select a representation for story events that maintains semantic meaning of textual story data while reducing sparsity of events.
The sparsity of events, in particular, results in poor story generation performance. 
Our experiments with different story representations during {\em event2event} generation back our hypothesis about event representation.
We found that the events that abstract away from natural language text the most improve the generative ability of a recurrent neural network story generation process. 
Event bigrams did not significantly harm the generative model and will likely help with coherence as they incorporate more history into the process, although story coherence is difficult to measure and was not evaluated in our experiments.

Although generalization of events away from natural language appears to help with event successor generation, it poses the problem of making story content unreadable. 
We introduced a second neural network, {\em event2sentence}, that learns to translate events with generalized or original words back into natural language.
This is important because it is possible for {\em event2event} to generate events that have never occurred (or have occurred rarely) in a story training corpus.
We maintain that being able to recover human-readable sentences from generalized events is valuable since our {\em event2event} experiments show use that they are preferred, and it is necessary to be able to fill in specifics later for dynamic storytelling.
We present a proposed pipeline architecture for filling in missing details in automatically generated partially generalized sentences.

The pursuit of automated story generation is nearly as old as the field of artificial intelligence itself. 
Whereas prior efforts saw success with hand-authored domain knowledge, machine learning techniques and neural networks provide a path forward toward the vision of open story generation, the ability for a computational system to create stories about any conceivable topic without human intervention other than providing a comprehensive corpus of story texts.


\section{Acknowledgments}
{\footnotesize
This work is supported by DARPA W911NF-15-C-0246. The views, opinions, and/or conclusions contained in this paper are those of the author and should not be interpreted as representing the official views or policies, either expressed or implied of the DARPA or the DoD.
The authors would like to thank Murtaza Dhuliawala, Animesh Mehta, and Yuval Pinter for technical contributions.}

\bibliographystyle{aaai}
\bibliography{misc,riedl}

\end{document}